\documentclass{article}

\usepackage{microtype}
\usepackage{graphicx}
\usepackage{subfigure}
\usepackage{booktabs}
\usepackage{amsmath}
\usepackage{adjustbox}
\usepackage{multirow}
\usepackage{float}

    \usepackage[final]{neurips_2025}

\usepackage[utf8]{inputenc} 
\usepackage[T1]{fontenc}    
\usepackage{hyperref}       
\usepackage{url}            
\usepackage{booktabs}       
\usepackage{amsfonts}       
\usepackage{nicefrac}       
\usepackage{microtype}      
\usepackage{xcolor}         

\title{Hallucination Detection and Mitigation with Diffusion in Multi-Variate Time-Series Foundation Models}

\author{%
  Vijja Wichitwechkarn\thanks{Corresponding Author.} \\
  Department of Engineering\\
  University of Cambridge \\
  Cambridge CB2 1PZ \\
  \texttt{vw273@cam.ac.uk} \\
  \And
  Charles Fox \\
  School of Computer Science \\
  University of Lincoln \\
  Lincoln LN6 7TS \\
  \texttt{ChFox@lincoln.ac.uk} \\
    \And
  Ruchi Choudhary \\
  Department of Engineering\\
  University of Cambridge \\
  Cambridge CB2 1PZ \\
  \texttt{rc488@cam.ac.uk} \\
}

\begin{document}

\maketitle

\begin{abstract}
Foundation models for natural language processing have many coherent definitions of hallucination and methods for its detection and mitigation. However, analogous definitions and methods do not exist for multi-variate time-series (MVTS) foundation models. We propose new definitions for MVTS hallucination, along with new detection and mitigation methods using a diffusion model to estimate hallucination levels. We derive relational datasets from popular time-series datasets to benchmark these relational hallucination levels. Using these definitions and models, we find that open-source pre-trained MVTS imputation foundation models relationally hallucinate on average up to 59.5\% as much as a weak baseline. The proposed mitigation method reduces this by up to 47.7\% for these models. The definition and methods may improve adoption and safe usage of MVTS foundation models.
\end{abstract}

\section{Introduction}
\label{Introduction}

Foundation models (FMs) trained on large and diverse datasets, that can be prompted to perform many types of computation, have enjoyed rapid progress in Natural Language Processing (NLP). Examples include Llama \cite{touvron2023llama}, ChatGPT \cite{achiam2023gpt}, Claude \cite{min2023recent} and Gemini \cite{anil2023gemini}. Models with similar capabilities are now also seen in other domains including time-series modelling. Recent works have shown that pre-trained models for time-series forecasting can be used effectively on unseen forecasting domains in a zero-shot manner. This is achieved by training on large quantities of time-series data from diverse domains as in Chronos \cite{ansari2024chronos}, TimesFM \cite{das2023decoder}, LagLlama \cite{rasul2023lag}, TimeGPT \cite{garza2023timegpt}, MOIRAI \cite{woo2024unifiedtraininguniversaltime}. Similar models have also been successful for time-series imputation such as MOMENT \cite{goswami2024moment}, TIMER \cite{liutimer}, TOTEM \cite{talukder2024totem}, TimesNet \cite{wu2022timesnet} and GPT4TS \cite{zhou2023one}.

We argue that pre-trained models for multi-variate time-series (MVTS) imputation are closer than MVTS forecasting to what are typically referred to as FMs in NLP as these can be prompted to handle different tasks. Prompts are the provided values and responses are the imputed values. For example, forecasting can be prompted for by masking future time-steps and asking the model to fill in these masked values; while interpolation can be prompted for with data from both before and after the missing period. Imputation therefore provides an interface for arbitrary question answering in MVTS. This work will therefore focus on these models, particularly MOMENT \cite{goswami2024moment} and TIMER \cite{liutimer} which are the only FMs of this type that currently have open-source weights available.

For MVTS question answering to be useful in real-world cases, a measure of confidence in the model's response is required, analogous to hallucination detection in NLP. To our knowledge, there is no literature on hallucination definition and detection in MVTS imputation, even with the advent of MVTS FMs. This is in stark contrast to NLP which has a large and active literature on defining and detecting different types of hallucination \cite{rawte2023survey, zhang2023siren, ye2023cognitive}. We here aim to bring MVTS hallucination detection closer to parity with NLP.

The following example highlights the utility of hallucination detection for MVTS question answering. Consider a MVTS system comprising three variables: temperature, vapour pressure deficit (VPD) and plant biomass. A farm operator may want to optimise the system to increase plant growth rate. To prompt for this question, an above average plant growth curve is provided and the model imputes the temperature and VPD it believes will result in this growth curve. This response, however, cannot be used since there is no measure of whether this response should be trusted. An inherently out of distribution question will inherently result in an out of distribution response. However, the response can still be correct. In order to use the response, MVTS hallucination detection is required.

The contributions of this work include the definition of two types of hallucination in the context of MVTS imputation: distributional and relational. These are defined using analogous definitions from the NLP literature. We use diffusion models \cite{ho2020denoising} for MVTS imputation and propose a method to detect and mitigate hallucination in its response. We also show that MVTS FMs hallucinate heavily using popular MVTS datasets, and that this can be detected and mitigated using our proposed methods. Source code will be provided here upon acceptance.


\subsection{Diffusion Model Preliminaries}

Diffusion models are probabilistic generative models that iteratively degrade data by introducing noise, then learn to reverse this process. This allows them to iteratively generate new samples by sampling from a simple prior, which is typically a Gaussian distribution \cite{yang2023diffusion}. They have become well known in image generation \cite{rombach2022high} and have been applied extensively to various fields including time-series generation \cite{yuan2024diffusion}, forecasting \cite{meijer2024rise} and imputation \cite{wang2024deep, yang2024survey}. Ever since diffusion models have been applied to time-series imputation \cite{tashiro2021csdi}, there has been growing work to improve them for this use case. These include improvements to the masking criteria during training \cite{xiao2023imputation, chen2023imdiffusion, liu2023pristi}, the architectures used \cite{alcaraz2022diffusion} and the sampling process \cite{wang2023observed}. Diffusion models have since become widely popular, becoming one of the best performing methods for time-series imputation \cite{zhou2024mtsci}. The background on diffusion models that is directly used in this work will be explained in the following sections. This includes the mathematical notations for Denoising Diffusion Probabilistic Models (DDPM) \cite{ho2020denoising} and conditioning through RePaint \cite{lugmayr2022repaint}.

\subsubsection{Unconditional Diffusion Models} 
In the forward process, samples from the training data $x_{0}$ are increasingly corrupted through the addition of Gaussian noise for $T$ time-steps to generate noisy samples $x_{1}, ..., x_{T}$:
\begin{equation}
q(x_{t}|x_{t-1}) := \mathcal{N}\left(x_{t}; \sqrt{1-\beta_{t}}x_{t-1, }\beta_{t}\mathbf{I}\right)
\end{equation}
\begin{equation}
    q(x_{1:T}|x_{0}) := \prod_{t=1}^{T}q(x_{t}|x_{t-1}).
\end{equation}
The Gaussian noise is determined by the variance schedule $\beta_{1}, ... , \beta_{T}$ which is typically linearly increasing. The forward process also admits sampling timestep $t$ directly:
\begin{equation}
    q(x_{t}|x_{0}) = \mathcal{N}\left(x_{t}; \sqrt{\bar{\alpha}_{t}}x_{0}, (1- \bar{\alpha}_{t})\mathbf{I}\right),
    \label{eq:direct_forward_dm}
\end{equation}
where $\alpha_{t}:= 1 - \beta_{t}$ and $\bar{\alpha}_{t}:=\prod^{t}_{s=1}\alpha_{s}.$

The reverse process is used to successively denoise the corrupted data by learning $p_{\theta}(x_{t-1}|x_{t})$ using a neural network with learnable parameters $\theta$:
\begin{equation}
    \mu_{\theta}(x_{t}, t) = \frac{1}{\sqrt{\alpha_{t}}}\left(x_{t} - \frac{1-\alpha_{t}}{\sqrt{1-\bar{\alpha}_{t}}}\epsilon_{\theta}(x_{t}, t)\right),
    \label{eq:mu}
\end{equation}
\begin{equation}
    p_{\theta}(x_{t-1}|x_{t}) := \mathcal{N}\left(x_{t-1}; \mu_{\theta}(x_{t}, t), \Sigma(x_{t}, t)\right),
    \label{eq:reverse_step_dm}
\end{equation}
\begin{equation}
    p_{\theta}(x_{0:T}) := p(x_{T})\prod_{t=1}^{T}p_{\theta}(x_{t-1}|x_{t}),
\end{equation}
where $\mu_{\theta}(x_{t}, t)$ is the predicted mean used to sample $x_{t-1}$ and $\epsilon_{\theta}(x_{t}, t)$ is the noise predicted by a neural network at time-step $t$. The original DDPM \cite{ho2020denoising} sets $\Sigma_{\theta}(x_{t}, t) = \sigma_{t}^{2}\mathbf{I}$, where $\sigma_{t}^{2} = \beta_{t}$. We will use this DDPM formulation of diffusion models in this work due to its popularity and simplicity.  


\subsubsection{Conditioning Diffusion with RePaint}\label{section:conditioning}
Diffusion models as described above are referred to as unconditional diffusion models as they do not directly allow for conditioning to be applied. It is however possible to guide the unconditional diffusion model using the RePaint \cite{lugmayr2022repaint} method. Here, components of the input vector $x$ are split into conditioning values $x^{(c)}$ and missing values $x^{(m)}$ to be imputed by the model. The missing values are sampled in the same way as Eq. \ref{eq:reverse_step_dm}:
\begin{equation}
    p_{\theta}(x_{t-1}^{(m)}|x_{t}^{(m)}) := \mathcal{N}\left(x_{t-1}^{(m)}; \mu_{\theta}(x_{t}^{(m)}, t), \Sigma(x_{t}^{(m)}, t)\right).
\end{equation}
The conditioning values however uses the corrupted version obtained using Eq. \ref{eq:direct_forward_dm}:
\begin{equation}
    q(x_{t}^{(c)}|x_{0}^{(c)}) = \mathcal{N}\left(x_{t}^{(c)}; \sqrt{\bar{\alpha}_{t}}x_{0}^{(c)}, (1- \bar{\alpha}_{t})\mathbf{I}\right).
    \label{eq:repaint_x0_to_xt}
\end{equation}
In this way, at each time-step $t$ of the reverse process, $x_{t}$ is composed of the imputed missing values $x_{t}^{(m)}$ obtained by denoising, and the conditioning values $x_{t}^{(c)}$ obtained by corrupting the actual given values to the correct noise level associated with the time-step $t$. The predicted mean at each diffusion time-step is then computed as usual using Eq. \ref{eq:mu}. 

We will use RePaint to condition an unconditional diffusion model trained on our dataset and impute missing values. This allows for arbitrary question answering using the prompt-response framework at inference without modification to the training procedure of the diffusion model.

\subsection{Hallucination Detection and Mitigation in NLP}

Example methods for hallucination detection and mitigation in NLP include the use of external knowledge retrieved from the web or task-specific databases to identify and correct non-factual content in responses \cite{peng2023check, shuster2021retrieval, lewis2020retrieval, chen2023purr, varshney2023stitch}. However, effective knowledge retrieval can be challenging and costly to run in practice \cite{mundler2023self}. It is unclear how this transfers to MVTS as there are no clear-cut facts to retrieve. 
There are also methods that do not use external knowledge but instead uses multiple samples from the same prompt to measure consistency of the generated information \cite{manakul2023selfcheckgpt, elaraby2023halo, zhang2023sac, farquhar2024detecting}. This can also be done using an ensemble of models \cite{du2023improving}. Similarly to these methods, this work will mitigate hallucination through sampling. The concept of consistency, however, does not transfer to MVTS as these require clear-cut facts and contradictions.
A separate hallucination detection model can also be trained to detect hallucination from the generated text \cite{chen2023hallucination, pacchiardi2023catch, mishra2024fine, zha2023alignscore} or the model's internal states \cite{su2024unsupervised}. This is the approach that will be adopted in this work using a diffusion model. There has also been work on scaling the generation of datasets that can be used to train these models \cite{su2024unsupervised, gu2024anah}. 
Hallucination mitigation can also be achieved through direct supervised finetuning \cite{gu2025mask, tian2023fine, lin2024flame, zhang2024self, chen2024grath}. However, the fine-tuned model still has to be used in conjunction with hallucination detection methods since they can still hallucinate, albeit at a potentially reduced rate.

\section{Defining Hallucination for Multi-Variate Time-Series Imputation} 

There is a large and active literature on defining, detecting and mitigating hallucination in NLP. In this context, hallucination is commonly defined as the behaviour when models generate responses with information that is {\em false} \cite{rawte2023survey, zhang2023siren, ye2023cognitive}. In time-series however, there are no clear-cut facts as in language. Consequently, there is no absolute truth to time-series, only what is probable relative to the provided context dataset. We therefore define \textbf{distributional hallucination} as a type of hallucination in time-series where the combination of the prompt and the generated response is out of distribution (OOD) with respect to a target dataset. Note that if an OOD prompt is provided to a model, all responses will automatically be classified as a distributional hallucination. This is important in the context of foundation models trained on large quantities of data since it is typically unknown whether a prompt is OOD or not. In practice, distributional hallucination is a continuous concept, so a threshold must be chosen to define a prompt-response pair as distributionally hallucinating.

Another definition of hallucination in NLP is the generation of {\em self-contradictory} responses \cite{mundler2023self};  with incoherent explanation and reasoning \cite{zhang2023language}; or responses that are irrelevant to the prompt \cite{gallifant2024peer}. These definitions will be used as the NLP analogue of what will be referred to as \textbf{relational hallucination}. A relation between a set of $N$ variables $\mathbf{x} = \{x_{1}, x_{2}, ..., x_{N}\}$ can be written as $f(\mathbf{x}) = 0$, where $f$ is some ground-truth function that defines the relation. The `relational error' which measures the degree of which the relation is broken can then be defined as $E_{r} := |f(\mathbf{x})|$. Relational hallucination can then be defined as the case when the model returns a set of variables that has `high' relational error, relative to some threshold. This occurs when the prompt and the response are incompatible, given $f$. This is analogous to a response that is irrelevant to the prompt in the NLP case. Additionally, relational hallucination can occur when the variables returned in the response are incompatible with themselves. This case is analogous to self-contradiction in NLP hallucination. Incoherent explanations and reasoning can also be seen as a form of self-contradiction. In the same way as distributional hallucination, relational hallucination is also defined relative to a given dataset.

\paragraph{Examples} In contrast to distributional hallucination, an OOD prompt may not necessarily result in relational hallucination. As a concrete example, consider the following case with three variables $\{x_{0}, x_{1}, x_{2}\}$ where the ground truth relation is addition: $x_{0} + x_{1} - x_{2} = 0$. The training dataset consists of $x_{0}, x_{1} \in [0, 10]$ and $x_{2} \in [0, 20]$. The combination of the prompt, where $x_{0}=21$ and $x_{1}=22$, and the response $x_{2}=43$, will be classified as distributionally hallucinating but not relationally hallucinating. The combination of the prompt, where $x_{0}=1$ and $x_{1}=2$, and the response $x_{2}=7$, will be classified as both distributionally hallucinating and relationally hallucinating. In this sense, relational hallucination is a subset of distributional hallucination.

\paragraph{Relational Hallucination is More Important} Distributional hallucination is important for detecting whether a question is OOD, which is typically not known at inference. Relational hallucination also captures all the in-distribution data, since by definition they all have correct relations between the variables. They however also extend to regions of the state space that is OOD. In this sense, relational hallucination is less restricted than distributional hallucination. Models being able to generalise to and operate in regions which are OOD is important as a large family of important question types are OOD. For example, to optimise variables to achieve better performances given the current data or to simulate a system under new conditions. This work will therefore focus on relational hallucination with the following proposed methods only applying to relational hallucination and not distributional hallucination.


\paragraph{Related Concepts}
{\em OOD detection} aims to detect test samples that do not exist in the training distribution \cite{yang2024generalized}. This is what we refer to as distributional hallucination in our work. We use the term hallucination as this is the common term used with respect to FMs in NLP. {\em Anomaly Detection} in contrast aims to detect unusual cases which may exist in the training set \cite{zamanzadeh2024deep}, assuming that the majority of training data is from the `correct' distribution and a minority of data is from an `anomalous' distribution. Anomaly detection can therefore be seen as OOD detection but with the definition of being `in distribution' replaced with being in the `correct distribution'. Anomaly detection in MVTS predicts which time indices within a single MVTS window correspond to anomalous values. Relational hallucination differs from these definitions, as it measures the compatibility of all the values in a MVTS window. A MVTS window can be out of distribution but still be relationally correct.



% ...



\section{Relational Hallucination Detection and Mitigation using Diffusion Models}
\label{section:our_methods}

Previous works have shown that diffusion models trained to generate images can detect hallucinations in their generated outputs \cite{aithal2024understanding}. They have also been successfully applied to MVTS imputation 
\cite{zhou2024mtsci} and anomaly detection \cite{chen2023imdiffusion}. We therefore consider diffusion models a promising candidate for arbitrary MVTS question answering through imputation, and the detection of relational hallucination.







\paragraph{Notations} To describe the prompt-response framework for MVTS imputation, we will use the following notation for each data point: $x_{i}$, where $i \in \mathcal{I}$ indexes the data dimension. A prompt is defined by specifying the set of variables that will be used as the prompt $i\in \mathcal{I}_{\mathrm{p}}$ and setting their values accordingly. The values for the remaining indices  $\mathcal{I}_{\mathrm{r}} = \mathcal{I} \setminus \mathcal{I}_{\mathrm{p}}$ will be masked and imputed by the model to generate the response. As before, the predicted mean at each diffusion time-step will be denoted $\mu_{i, t}$ where $t \in \{0, ..., T\}$. Note that denoising decrements the time-step from $T$ to $0$. The final output (prediction) from the model ($t=0$) will be denoted as $\hat{x}_{i}$, where the imputed response is $\hat{x}_{i}, \forall i\in\mathcal{I}_{\mathrm{r}}$, and $\hat{x}_{i} \approx x_{i}, \forall i\in\mathcal{I}_{\mathrm{p}}$. 

\paragraph{Conditioning} Once the prompt is defined, RePaint \cite{lugmayr2022repaint} is used to condition an unconditional diffusion model trained on the dataset. The prompt is used as the conditioning $x^{(c)}$ in RePaint  as described in Section \ref{section:conditioning}. This allows diffusion models to act as a prompt-response model for general time-series question answering.

\begin{figure*}
    \centering
    \includegraphics[width=1.0\linewidth, trim=0 50 0 0, clip]{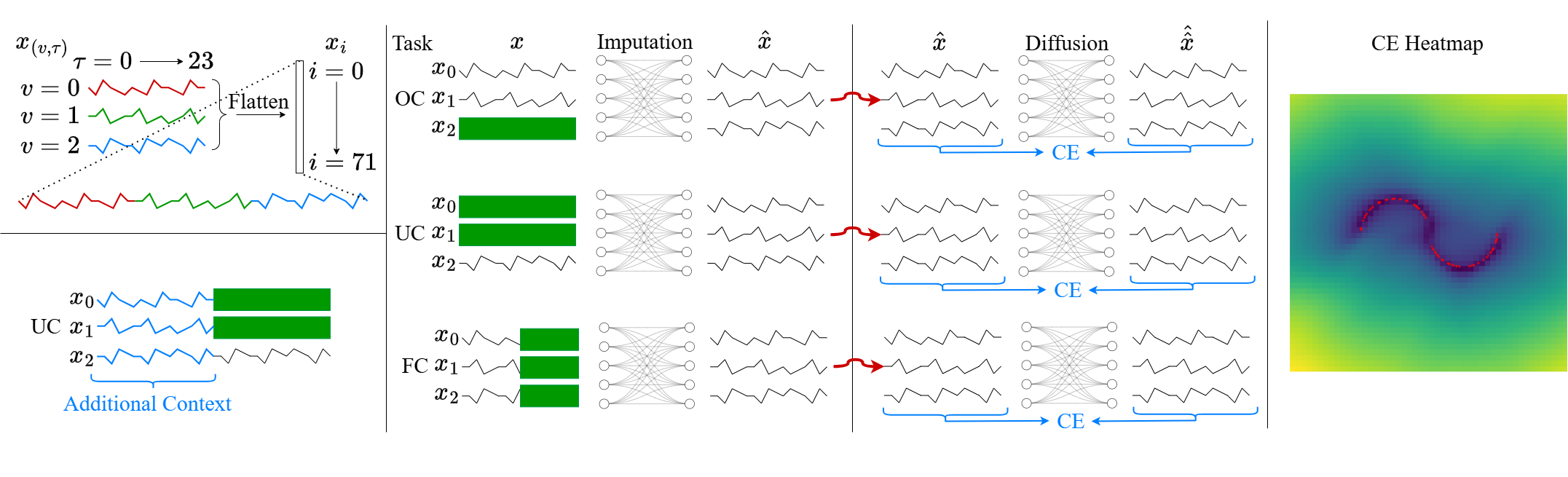}
    \caption{(\textbf{Left Top}) Index decomposition of a data point. \textbf{(Left Bottom)} Example of additional context provided on the UC task. \textbf{(Middle Left)} Schematic showing the different type of tasks (OC, UC and FC) for the prompt. Masked variables are shown as blank green boxes and unmasked variables are used as the prompt. The imputation process can be done using the diffusion model or pre-trained foundation models. \textbf{(Middle Right)} The combination of the prompt and response obtained from the imputation is used as the prompt for a diffusion process, which is used to compute Combined Error (CE) metric.
    (\textbf{Right}) CE heatmap for a diffusion model fit to a small non-linear 2D dataset. Red dots are the data points in the training set and darker colours correspond to lower CE values.}
    \label{fig:overview}
\end{figure*}

\paragraph{Relational Hallucination Metric}
 We propose the Combined Error (CE) metric that can be used to estimate the level of relational hallucination and a method to extract it from a diffusion model trained on the dataset. This metric can be computed for a given prompt-response pair $\hat{x}_{i},  \forall i \in \mathcal{I}$ obtained from some model such as a foundation model. It is computed by using RePaint to condition the diffusion model and setting the prompt as $x_{i} := \hat{x}_{i}, \forall i \in \mathcal{I}$. The output of this process will be referred to as $\hat{\hat{x}}_{i}$ where the double hat denotes a prediction where the target is a previous prediction. The CE metric can be computed as
\begin{equation}
    M_{\mathrm{CE}} = \mathrm{RMSE}_i(\hat{\hat{x}}_{i}, \hat{x}_{i} ), \;\forall i \in \mathcal{I},
    \label{eqn:def-ce}
\end{equation}
where the root mean square error is taken across the data dimension. Note that $\hat{\hat{x}}_i$ can be computed using a single denoising step (the final time-step going $t=0$). This is because RePaint allows the diffusion process to be skipped to the final step for all conditioning values, which in this case is the entire data dimension. This is done using the forward process (Eq. \ref{eq:repaint_x0_to_xt}). Denoising using Eq. \ref{eq:mu} is therefore only done on the final diffusion time-step and obtaining this metric is not computationally expensive. This process is shown in Fig. \ref{fig:overview} (middle right). 

To highlight properties of the CE metric, a diffusion model was trained on a small nonlinear 2D dataset. Fig. \ref{fig:overview} (right) visualises the value of the CE metric for each point in this space. The CE metric is low in regions where the relations hold. This is true even in OOD regions without data. We also tested variations of metrics similar to \cite{aithal2024understanding}. However, these are not effective, as shown in Appendix \ref{appendix:othermetrics}.



\paragraph{Hallucination Detection}
To use the proposed CE metric to gauge the expected level of relational hallucination, a dataset-specific scale is required to determine whether the metric is high/low relative to the dataset. We propose a simple method. Firstly, the CE metric is obtained for all the prompt-response pairs obtained from the training set, over all the imputation tasks. The quartiles of the CE metric are then computed. These quartiles are then used to classify the prompt-response pairs into classes of expected relational hallucination levels at inference: low (below the second quartile), medium (between the second and third quartile) and high (greater than the third quartile).

\paragraph{Hallucination Mitigation}
we also propose a simple method for mitigating relational hallucination for non-deterministic models. For a given prompt, $N$ responses are sampled from the model. The CE metric can then be computed for all the obtained prompt-response pairs $\left(\hat{x}^{(j)}, M_{CE}^{(j)}\right) \in \mathcal{X}_{\text{sampled}}$. The prompt-response pair with the lowest metric $\hat{x}^{(j^{*})}$, where $j^{*} = \arg\min \left(M_{CE}^{(j)}\right)$, is then selected as the response with the expected lowest relational hallucination.

\section{Experiments} 
\label{section:experiments}

\paragraph{Why We Need New Datasets} In real-world settings, the ground-truth relation function $f$ is typically unknown. So, the ground-truth relational error cannot be computed for a given prompt-response pair. This work proposes the use of diffusion models to generate a metric for a given prompt-response pair, that can be used as an indirect measure for the relational error, and hence relational hallucination. To evaluate our proposed methods however, we require a dataset with a known $f$ that can be used to compute the ground-truth relational error. We introduce five such datasets: the relational electricity consuming load (rECL), relational weather (rWTH), relational traffic (rTraffic), relational illness (rIllness) and relational electricity transformer temperature (rETT) dataset, which are relational variations of popular MVTS datasets. We aim to demonstrate that our method can estimate relational hallucination by applying it to datasets where $f$ is known and hence the ground-truth relational error can be computed.


\paragraph{Datasets}
The rECL dataset is derived from the electricity consuming load dataset \cite{ecldataset}. In this case, the first and second variable are the electric consumption data for two different clients. The third variable is generated by taking the difference between the first two variables. The rWTH dataset is derived from the weather dataset \cite{wthdataset}, another commonly used time-series dataset. Temperature $T$ and humidity $H$ is taken from this dataset and a third variable, vapour pressure deficit (VPD) is created which is a non-linear function of temperature and humidity $f_{vpd}(T, H) = 0.6108 \times \exp \left( \frac{17.27 \times T}{T + 237.3} \right) \times \left(1 - H\right),$ where $T$ is the temperature in Celsius and $H$ is the relative humidity expressed as a decimal. This is a real-world example dataset that include the variables important for agriculture. rTraffic is derived from the traffic dataset \cite{trafficdataset}, where the first two variables are the same as the original and the third variable is the sum of the two. rIllness is derived from the Illness dataset \cite{illnessdataset}, where the first two variables are the same as the original and the third variable is the difference between the two. rETT is derived from the ETTh1 dataset \cite{zhou2021informer}, where the first two variables are the `middle useful load' and the oil temperature, and the third variable is the product of the two. A context length of $L=24$ is used for each data point, this results in each data point $x_{i}$ with $i \in \{0, ..., 71\}$ since there are three variables. A schematic of this is shown in Fig. \ref{fig:overview} (left top).


\paragraph{Tasks}

We consider three types of prompts in this work, which will be referred to as tasks. To describe these tasks, we will split the indices of each data point $x_{i}$ using the following notation $x_{(v, \tau)}$ where $v\in \{0, 1, 2\}$ indexes the variables and $\tau\in \{0, ..., 23\}$ indexes the time-step within the data point. This is not the diffusion time-step $t$ and this splitting of the indices is only used in the description of the three tasks below. Illustrations of this index decomposition and the tasks are shown in the top left and middle left of Fig.~\ref{fig:overview}.

\begin{itemize}
    \item \textbf{Over-constrained (OC)} - this task include prompts that constrain the response such that there is one correct response. For this task the indices $v\in \{0, 1\}$ and $\tau \in \{0, ..., 23\}$ will be used for the prompt. 
    \item \textbf{Under-constrained (UC)} - this task include prompts that under-constrain the response such that there are multiple correct responses. For this task the indices $v\in \{2\}$ and $\tau \in \{0, ..., 23\}$ will be used for the prompt.
    \item \textbf{Forecast (FC)} - this task does not separately constrain the related variables but provide the past variables as the prompt. For this task the indices $v\in \{0, 1, 2\}$ $\tau \in \{0, ..., 11\}$ will be used for the prompt.
\end{itemize}

\begin{table*}[h]
\begin{center}
\begin{small}
\begin{sc}
\begin{adjustbox}{width=\textwidth}

\begin{tabular}{ll|cc|cc|cc}
\toprule
\multirow{2}{*}{Dataset} & \multirow{2}{*}{Model} & \multicolumn{2}{|c|}{Task = OC} & \multicolumn{2}{|c}{Task = UC} &  \multicolumn{2}{|c}{Task = FC} \\
 &  & $E_{r}$ & $\langle E_{r}\rangle / \langle E_{r}\rangle ^{\text{(baseline)}}$ & $E_{r}$ & $\langle E_{r}\rangle / \langle E_{r}\rangle ^{\text{(baseline)}}$ & $E_{r}$ &$\langle E_{r}\rangle / \langle E_{r}\rangle ^{\text{(baseline)}}$ \\
\midrule

\multirow{3}{*}{rECL} & Baseline & $  0.9841 \pm 0.3500  $ & $  1.0000  $ & $  0.9841 \pm 0.3500  $ & $  1.0000  $ & $  0.9841 \pm 0.3500  $ & $  1.0000  $  \\
& DM (ours) & $  \mathbf{0.1491 \pm 0.0543}  $ & $  0.1515  $ & $  \mathbf{0.0572 \pm 0.0265}  $ & $  0.05812  $ & $  \mathbf{0.0138 \pm 0.0037}  $  & $ 0.0140 $ \\
& MOMENT & $  0.5744 \pm 0.2019  $ & $  0.5837  $ & $  0.5495 \pm 0.2203  $ & $  0.5584  $ & $  0.2164 \pm 0.1272  $  & $0.2199$ \\
& TIMER & $  0.6197 \pm 0.2400  $ & $  0.6297  $ & $  0.5121 \pm 0.2127  $ & $  0.5203  $ & $  0.2182 \pm 0.1260  $  & $0.2217$ \\
\midrule

\multirow{3}{*}{rWTH} & Baseline & $0.6283 \pm 0.3026$ & $1.0000$ & $0.6283 \pm 0.3026$ & $1.0000$ & $0.6283 \pm 0.3026$ & $1.0000$ \\
& DM (ours) & $  \mathbf{0.0550 \pm 0.0600}  $ & $  0.0875  $ & $  \mathbf{0.0932 \pm 0.0673}  $ & $  0.1483  $ & $  \mathbf{0.0160 \pm 0.0076}  $  & $0.0255$ \\
& MOMENT & $  0.2683 \pm 0.1820  $ & $  0.4270  $ & $  0.2651 \pm 0.1801  $ & $  0.4219  $ & $  0.0785 \pm 0.0507  $ &  $0.1249$ \\
& TIMER & $  0.6477 \pm 0.2167  $ & $  1.0309 $ & $  0.3492 \pm 0.2179  $ & $  0.5558 $ & $  0.2459 \pm 0.0467  $ &  $0.3913$ \\
\midrule

\multirow{3}{*}{rTraffic} & Baseline & $0.1058 \pm 0.0579$ & $1.0000$ & $0.1058 \pm 0.0579$ & $1.0000$ & $0.1058 \pm 0.0579$ & $1.0000$ \\
& DM (ours) & $  \mathbf{0.0027 \pm 0.0010}  $ & $  0.0255  $ & $  \mathbf{0.0096 \pm 0.0056}  $ & $  0.0907  $ & $  \mathbf{0.0014 \pm 0.0006}  $ &  $  0.0132  $  \\
& MOMENT & $  0.0513 \pm 0.0314  $ & $  0.4849  $ & $  0.0533 \pm 0.0326  $ & $  0.5038  $ & $  0.0046 \pm 0.0040  $ &  $  0.0435  $  \\
& TIMER & $  0.0974 \pm 0.0284  $ & $  0.9206  $ & $  0.1006 \pm 0.0309  $ & $  0.9509  $ & $  0.0043 \pm 0.0035  $ &  $  0.0409 $  \\
\midrule

\multirow{3}{*}{rIllness} & Baseline & $4469 \pm 3585$ & $1.0000$ & $4469 \pm 3585$ & $1.0000$ & $4469 \pm 3585$ & $1.0000$ \\
& DM (ours) & $  \mathbf{1521 \pm 951.6}  $ & $  0.3403  $ & $  \mathbf{996.4 \pm 661.9}  $ & $  0.2230  $ & $  \mathbf{380.1 \pm 224.7}  $ &  $  0.0851  $  \\
& MOMENT & $  3183 \pm 1913  $ & $  0.7122  $ & $  3815 \pm 2098  $ & $  0.8537  $ & $  1174 \pm 681.0  $ &  $   0.2627 $  \\
& TIMER & $  3314 \pm 1545  $ & $   0.7416 $ & $  3459 \pm 2096  $ & $   0.7740 $ & $  1554 \pm 1384  $ &  $  0.3477  $  \\
\midrule

\multirow{3}{*}{rETT} & Baseline & $  0.5600 \pm 0.2894  $ & $  1.0000  $ & $  0.5600 \pm 0.2894  $ & $  1.0000  $ & $  0.5600 \pm 0.2894  $ & $  1.0000  $ \\
& DM (ours) & $  \mathbf{0.2312 \pm 0.1704}  $ & $  0.4129  $ & $  \mathbf{0.2875 \pm 0.1750}  $ & $  0.5134  $ & $  \mathbf{0.0597 \pm 0.0398}  $ &  $  0.1066  $  \\
& MOMENT & $  \mathbf{0.3796 \pm 0.2392}  $ & $  0.6779  $ & $  \mathbf{0.3231 \pm 0.1977}  $ & $  0.5770  $ & $  0.1440 \pm 0.0908  $ &  $  0.2571  $  \\
& TIMER & $  \mathbf{0.3177 \pm 0.2185}  $ & $  0.5673  $ & $  0.4666 \pm 0.2721  $ & $  0.8332  $ & $  0.2271 \pm 0.1324  $ &  $  0.4055  $  \\
\bottomrule

\end{tabular}

\end{adjustbox}
\end{sc}
\end{small}
\end{center}
\caption{Relational error $E_{r}$ for each model on each dataset (lower is better). The best values for each dataset are highlighted in bold. The mean values relative to the weak baseline are also given.}
\label{table:models_hallucinate}
\end{table*}

\paragraph{Models} that will be evaluated on the tasks above are:

\begin{itemize}
    \item \textbf{Baseline} - Since each dataset will have different scales, a baseline is required to compare against. A weak baseline that returns the training set mean for each variable for all responses will be used.

    \item \textbf{Diffusion Model} - The diffusion model trained on each dataset, which will be used for hallucination detection on that dataset. It can conveniently also be used for question answering. This will serve as a stronger baseline. The model uses a simple five layer MLP with  1 million parameters. 

    \item \textbf{MOMENT} \cite{goswami2024moment} - A MVTS foundation model using a transformer encoder architecture, pre-trained on Time-Series Pile (20GB). We use the large model with 24 layers and 385M parameters.  This model will be used for question answering only. MOMENT models MVTS in a channel-independent manner, a popular choice \cite{nie2022time}. As shown in Fig. \ref{fig:overview} (left bottom), we therefore provide additional context (24 time-steps) to each task to allow MOMENT to function on tasks like the OC and UC task. This makes the task easier.
    


    \item \textbf{TIMER} \cite{liutimer} - A MVTS foundation model using a transformer decoder architecture, pre-trained on the UTSD-4G dataset (1.2GB).  It has 4 layers and 2M parameters. This model will be used for question answering only. Since TIMER requires at least the first token (24 time-steps) to be provided, additional context is also provided in the same way as MOMENT. This allows for a fair comparison.

\end{itemize}

\paragraph{Implementation} is in Python 3.11 using PyTorch. The diffusion models trained were all MLPs with five hidden layers of size 512. A linear variance schedule ranging from a value of 1e-4 to 1e-2 was used with 1000 diffusion steps. Models were trained using the ADAM optimizer \cite{diederik2014adam}, one-cycle learning rate scheduler \cite{smith2019super}, a maximum learning rate of 1e-3 and batch size of 1024. All models were trained up to a maximum of 8000 epochs with early stopping. The model with best validation loss was used for all subsequent experiments. The relational datasets use all the data present in the original dataset and were split into train, validation and test sets with a ratio of 5:1:1 in a chronologically increasing manner such that there is no overlap in time. Training runs on a single NVIDIA T1000 in 2-22 hours depending on the dataset.

\subsection{Multi-Variate Time-Series Models Hallucinate}
\label{section:expt_mvts_hallu}

As the ground-truth relation $f$ is known for our datasets, the degree of relational hallucination exhibited by a model can be quantified. This is achieved by using each model to respond to all the prompts from the OC, UC and FC tasks on each dataset (test set), and then computing the relational error $E_{r}$. The lower the average $E_{r}$ is, the better. As each dataset has different value scales, all $E_{r}$ comparisons are relative to the weak baseline. Since the diffusion model was trained on the training set of each dataset, it can be taken as a strong baseline. These values are shown in Table~\ref{table:models_hallucinate} (mean and standard deviation) for each model, task and dataset (test set). The mean values normalised by the baseline's mean is also provided so that it is easier to compare across the datasets with different scales.


The results show that even with the handicap of being given extra context, both the pre-trained foundation models (MOMENT and TIMER) hallucinate heavily. They typically hallucinate less than the weak baseline but in some cases can match or even exceed it. The diffusion model (strong baseline) hallucinates the least, but nevertheless still hallucinates. All models relationally hallucinate the least on the FC task. This may be because there are no hard constraints on the values that must be predicted and the model is free to sample/predict values that are relationally correct. Averaging over the tasks and datasets, the relational hallucination level of the diffusion model, MOMENT and TIMER are 15.3\%, 44.6\% and 59.5\% the values of the weak baseline. The results demonstrate that even models trained on each dataset can relationally hallucinate relative to that dataset, with this being exhibited much strongly in pre-trained foundation models.

\subsection{Estimation of Hallucination Levels at Inference}
\label{section:expt_hallu_detect}

The following proposed quartile thresholding method is used to classify responses by their expected relational hallucination level: low, medium and high. This can be evaluated by computing the relational error $E_{r}$ for all the prompt-response pairs classified into each class. This gives us the distribution of $E_{r}$ for each class. The overlap coefficient between the distribution of $E_{r}$ for the low and high classes can be computed. These distributions will be referred to as $\mathbf{P^{\mathrm{(L)}}}$ and $\mathbf{P^{\mathrm{(H)}}}$, respectively. They are of the form $\mathbf{P} = \{P_{0}, P_{1}, ... , P_{n-1}\}$, where $n$ is the number of bins and the values are the probability in each bin with $\sum_{k}P_{k} = 1$. The overlap coefficient between them can be computed as $\sum_{k=0}^{n-1} \min\left(P^{(\mathrm{L})}_{k}, P^{(\mathrm{H})}_{k}\right).$ Lower coefficients mean better hallucination detection. A value of zero implies zero overlap, and a value of one implies that the distributions are identical. The results (mean and standard deviation) obtained for the models on each dataset averaged over five runs are shown in Table \ref{table:the_rest}. The overlap coefficients are low (generally below 1\%) except for the the rETT dataset which is a moderate value of around 15\%.

\begin{figure}[h]
    \centering
    \includegraphics[width=0.9\linewidth]{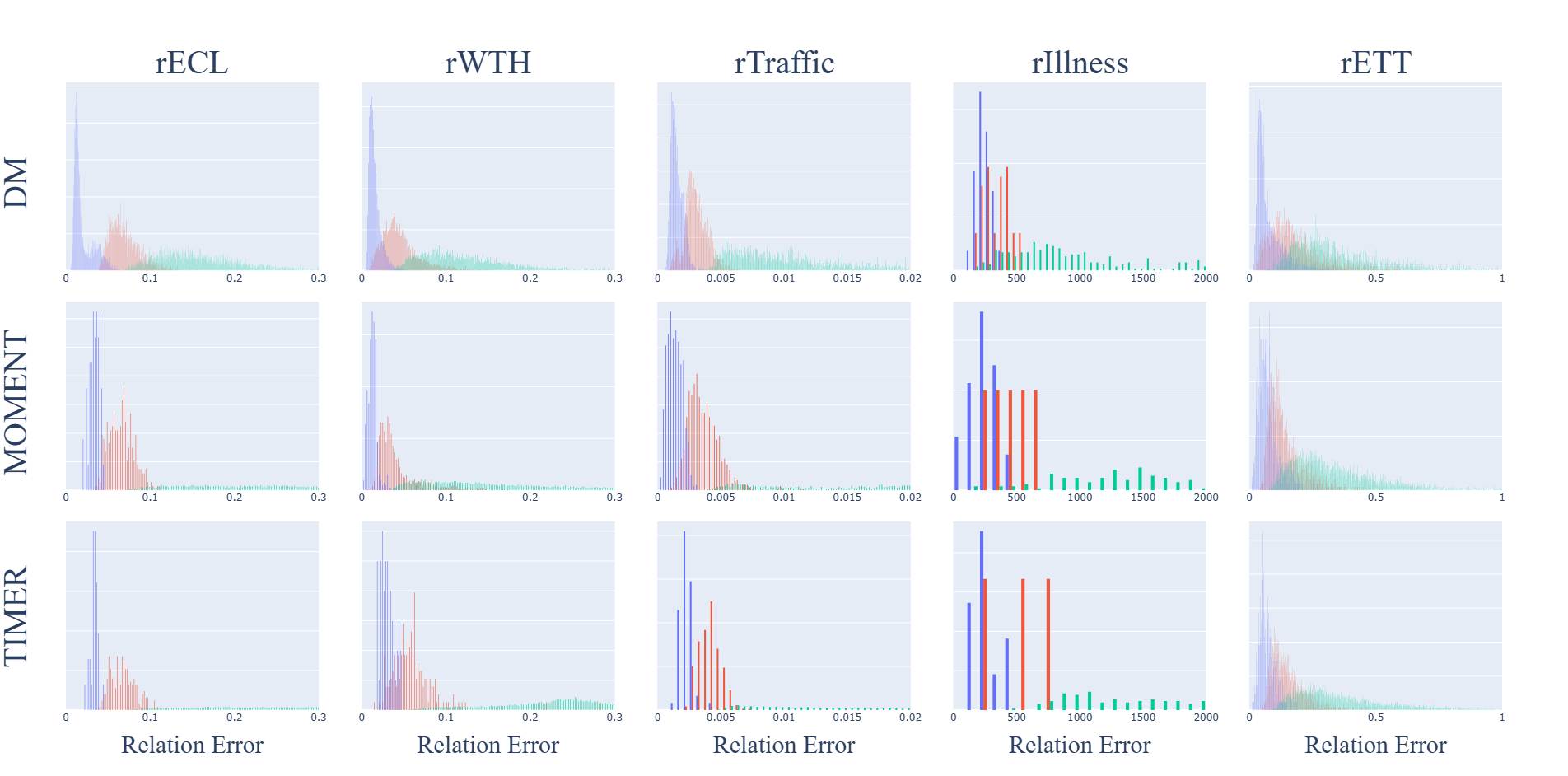}
    \caption{Histogram showing the distribution of relational error for the data points with expected low (blue), medium (red) and high (green)  hallucination level. The $x$-axis is the relational error and the $y$-axis is the probability. The subplots are aligned by dataset (column) and model (row).}
    \label{fig:quartile_hist}
\end{figure}

The histogram of the relational error for each class of hallucination level is shown in Fig. \ref{fig:quartile_hist}. The results show that quartile thresholding is a simple and effective way to classify responses into their expected relational hallucination levels where the distributions with high and low hallucination have low overlap.











\begin{table}[h]
\begin{center}
\begin{small}
\begin{sc}
\begin{adjustbox}{width=\columnwidth}
\begin{tabular}{llcccc}
\toprule
Dataset & Model & Overlap Coefficient & $\Delta_{E_{r}}$(OC) & $\Delta_{E_{r}}$ (UC) &  $\Delta_{E_{r}}$ (FC) \\
\midrule

\multirow{3}{*}{rECL} & DM (ours) &     $  0.0008 \pm 0.0003  $     & $  0.6230 \pm 0.1060 $ & $   0.4789 \pm 0.1045 $ & $  0.6705 \pm 0.1061 $\\
& MOMENT &   $0.0000 \pm 0.0000$       & $  0.7397 \pm 0.1159   $ & $   0.7549 \pm 0.12  $ & $  0.7249 \pm 0.2889   $\\
& TIMER &     $ 0.0000 \pm 0.0000   $     & $ 0.7289 \pm 0.1854  $ & $ 0.705 \pm 0.1919  $ & $  0.5468 \pm 0.2581 $\\
\midrule

\multirow{3}{*}{rWTH} & DM (ours) &     $  0.0167 \pm 0.0007  $     & $ 0.7051 \pm 0.1656  $ & $  0.5089 \pm 0.1782 $ & $ 0.8550 \pm 0.2287  $\\
& MOMENT &     $0.0005 \pm 0.0005$     & $   0.8086 \pm 0.1195  $ & $  0.8143 \pm 0.1242   $ & $  0.7231 \pm 0.239   $\\
& TIMER &    $ 0.0002 \pm 0.0001   $      & $  0.8175 \pm 0.149 $ & $  0.7877 \pm 0.1752 $ & $ 0.7753 \pm 0.1772  $\\
\midrule

\multirow{3}{*}{rTraffic} & DM (ours) &    $ 0.0009 \pm 0.0003   $      & $ 0.6600 \pm 0.0902  $ & $  0.4493 \pm 0.1162 $ & $  0.8057 \pm 0.2055 $\\
& MOMENT &    $ 0.0000 \pm 0.0000   $      & $  0.7769 \pm 0.1229   $ & $  0.7638 \pm 0.1197   $ & $   0.8739 \pm 0.1409 $\\
& TIMER &    $  0.0000 \pm 0.0000  $      & $ 0.7743 \pm 0.1679  $ & $  0.7985 \pm 0.161 $ & $ 0.5228 \pm 0.2608  $\\
\midrule

\multirow{3}{*}{rIllness} & DM (ours) &     $  0.0111 \pm 0.0040  $     & $  0.7311 \pm 0.1372 $ & $ 0.7079 \pm 0.1225  $ & $ 0.8787 \pm 0.1986  $\\
& MOMENT &   $  0.0008 \pm 0.0017  $       & $  0.8392 \pm 0.2418   $ & $  0.7860 \pm 0.1810   $ & $  0.9647 \pm 0.3061   $\\
& TIMER &     $  0.0000 \pm 0.0000  $     & $  0.6475 \pm 0.2664 $ & $ 0.5988 \pm 0.2501  $ & $ 0.5343 \pm 0.3847  $\\
\midrule

\multirow{3}{*}{rETT} & DM (ours) &     $  0.1538 \pm 0.0054  $     & $ 0.7681 \pm 0.2128  $ & $ 0.6724 \pm 0.2382  $ & $ 0.9074 \pm 0.3903  $\\
& MOMENT &    $  0.0880 \pm 0.0045  $      & $   0.8291 \pm 0.1645  $ & $  0.8269 \pm 0.1597   $ & $  0.7353 \pm 0.3114   $\\
& TIMER &      $  0.0371 \pm 0.0061  $    & $ 0.7752 \pm 0.2439  $ & $ 0.7833 \pm 0.1976  $ & $ 0.672 \pm 0.2444  $\\
\bottomrule

\end{tabular}
\end{adjustbox}
\end{sc}
\end{small}
\end{center}
\caption{
Overlap coefficient between the data distribution classified as low and high hallucination (lower is better). Relative change in relational error $\Delta_{E_{r}}$ of the selected response using filtering (lower is better).}
\label{table:the_rest}
\end{table}


\subsection{Mitigation of Hallucination at Inference}
\label{section:expt_hallu_mitigate}
The proposed filtering method for mitigating relational hallucination can be evaluated using the ground-truth relation $f$ that is available for the proposed datasets used in this work. This can be achieved by computing the relational error $E_{r}$ for the response selected by filtering $E_{r}^{(j^{*})}$ and comparing it to the mean relational error for the $N$ sampled responses $\langle E_{r} \rangle = \frac{1}{N}\sum_{j=1}^{N}E_{r}^{(j)}$. The relative change in relational error can then be computed $\Delta_{E_{r}} = E_{r}^{(j^{*})} / \langle E_{r} \rangle$. The lower $\Delta_{E_{r}}$ is the better, with $\Delta_{E_{r}}=1$ meaning there is no improvement. Since the foundation models (MOMENT and TIMER) are deterministic, a simple way make them non-deterministic and sample from them is to activate the dropout layers used for their training. The sample with the lowest CE is then selected. Instead of computing $\Delta_{E_{r}}$ relative to the mean of the ensemble, it should be relative to the response from the model with deactivated dropout.


The relative change in relational error $\Delta_{E_{r}}$ for each dataset averaged over 20 runs is given in Table \ref{table:the_rest} (mean and standard deviation). The average relative change $\Delta_{E_{r}}$ is always less than unity, which means that the filtering method is effective, even when the pre-trained foundation models with dropout. The proposed method can on average reduce the relational error by up to 55.0\% for the diffusion model and 47.7\% for the pre-trained foundation models. This demonstrates that filtering using CE is a simple and effective method for mitigating relational hallucination.

\section{Conclusion}
\label{section:conclusion}
Hallucination in MVTS imputation has been defined using analogies from established definitions in NLP. Pre-trained open-source MVTS foundation models are seen to hallucinate in this manner. By training a diffusion model on data in a target domain and extracting the proposed CE metric, it is possible to detect and mitigate MVTS hallucination, being able to on average reduce the hallucination of pre-trained FMs by up to 47.7\%. This work encourages the responsible use of MVTS FMs by formally defining, detecting and mitigation MVTS hallucination.

\paragraph{Limitations and Further Work}
While our work shows promising results, it is largely intuition-driven and empirical. For instance, our mitigation method statistically improves responses, but is not guaranteed to always do so. This can be improved through further work on methods with stronger theoretical guarantees. Additionally, the MLP architecture used for the diffusion model is simple, and hence does not naturally support variable length responses. Switching to other architectures such as the transformer will address this issue. Since we stack each variable into one time series as the input to the model, the simple MLP architecture does not scale well to a high number of variables or long windows. Exploring the use of latent diffusion or tokenisation may address this issue. There is, however, currently no consensus on the best architectures for MVTS and hence further work is required. Although the current method used to convert deterministic pre-trained MVTS foundation models into non-determinstic ones that can be sampled works, it is very simple. Exploring decoding strategies and methods from NLP for sampling responses that can be applied to MVTS is another promising direction.


\bibliography{example_paper}

\begin{thebibliography}{10}

\bibitem{touvron2023llama}
Hugo Touvron, Louis Martin, Kevin Stone, Peter Albert, Amjad Almahairi, Yasmine Babaei, Nikolay Bashlykov, Soumya Batra, Prajjwal Bhargava, Shruti Bhosale, et~al.
\newblock Llama 2: Open foundation and fine-tuned chat models.
\newblock {\em arXiv:2307.09288}, 2023.

\bibitem{achiam2023gpt}
Josh Achiam, Steven Adler, Sandhini Agarwal, Lama Ahmad, Ilge Akkaya, Florencia~Leoni Aleman, Diogo Almeida, Janko Altenschmidt, Sam Altman, Shyamal Anadkat, et~al.
\newblock {GPT}-4 technical report.
\newblock {\em arXiv:2303.08774}, 2023.

\bibitem{min2023recent}
Bonan Min, Hayley Ross, Elior Sulem, Amir Pouran~Ben Veyseh, Thien~Huu Nguyen, Oscar Sainz, Eneko Agirre, Ilana Heintz, and Dan Roth.
\newblock Recent advances in natural language processing via large pre-trained language models: A survey.
\newblock {\em ACM Computing Surveys}, 56(2):1--40, 2023.

\bibitem{anil2023gemini}
Rohan Anil, Sebastian Borgeaud, Yonghui Wu, Jean-Baptiste Alayrac, Jiahui Yu, Radu Soricut, Johan Schalkwyk, Andrew~M Dai, Anja Hauth, Katie Millican, et~al.
\newblock Gemini: A family of highly capable multimodal models.
\newblock {\em arXiv preprint arXiv:2312.11805}, 1, 2023.

\bibitem{ansari2024chronos}
Abdul~Fatir Ansari, Lorenzo Stella, Caner Turkmen, Xiyuan Zhang, Pedro Mercado, Huibin Shen, Oleksandr Shchur, Syama~Sundar Rangapuram, Sebastian~Pineda Arango, Shubham Kapoor, et~al.
\newblock Chronos: Learning the language of time series.
\newblock {\em arXiv:2403.07815}, 2024.

\bibitem{das2023decoder}
Abhimanyu Das, Weihao Kong, Rajat Sen, and Yichen Zhou.
\newblock A decoder-only foundation model for time-series forecasting.
\newblock {\em arXiv:2310.10688}, 2023.

\bibitem{rasul2023lag}
Kashif Rasul, Arjun Ashok, Andrew~Robert Williams, Arian Khorasani, George Adamopoulos, Rishika Bhagwatkar, Marin Bilo{\v{s}}, Hena Ghonia, Nadhir~Vincent Hassen, Anderson Schneider, et~al.
\newblock Lag-llama: Towards foundation models for time series forecasting.
\newblock {\em arXiv:2310.08278}, 2023.

\bibitem{garza2023timegpt}
Azul Garza and Max Mergenthaler-Canseco.
\newblock {TimeGPT-1}.
\newblock {\em arXiv:2310.03589}, 2023.

\bibitem{woo2024unifiedtraininguniversaltime}
Gerald Woo, Chenghao Liu, Akshat Kumar, Caiming Xiong, Silvio Savarese, and Doyen Sahoo.
\newblock Unified training of universal time series forecasting transformers, 2024.

\bibitem{goswami2024moment}
Mononito Goswami, Konrad Szafer, Arjun Choudhry, Yifu Cai, Shuo Li, and Artur Dubrawski.
\newblock Moment: A family of open time-series foundation models.
\newblock {\em arXiv:2402.03885}, 2024.

\bibitem{liutimer}
Yong Liu, Haoran Zhang, Chenyu Li, Xiangdong Huang, Jianmin Wang, and Mingsheng Long.
\newblock Timer: Generative pre-trained transformers are large time series models.
\newblock In {\em Forty-first International Conference on Machine Learning}, 2024.

\bibitem{talukder2024totem}
Sabera Talukder, Yisong Yue, and Georgia Gkioxari.
\newblock Totem: Tokenized time series embeddings for general time series analysis.
\newblock {\em arXiv:2402.16412}, 2024.

\bibitem{wu2022timesnet}
Haixu Wu, Tengge Hu, Yong Liu, Hang Zhou, Jianmin Wang, and Mingsheng Long.
\newblock Timesnet: Temporal 2{D}-variation modeling for general time series analysis.
\newblock {\em arXiv:2210.02186}, 2022.

\bibitem{zhou2023one}
Tian Zhou, Peisong Niu, Liang Sun, Rong Jin, et~al.
\newblock One fits all: Power general time series analysis by pretrained {LM}.
\newblock {\em Advances in neural information processing systems}, 2023.

\bibitem{rawte2023survey}
Vipula Rawte, Amit Sheth, and Amitava Das.
\newblock A survey of hallucination in large foundation models.
\newblock {\em arXiv:2309.05922}, 2023.

\bibitem{zhang2023siren}
Yue Zhang, Yafu Li, Leyang Cui, Deng Cai, Lemao Liu, Tingchen Fu, Xinting Huang, Enbo Zhao, Yu~Zhang, Yulong Chen, et~al.
\newblock Siren's song in the {AI} ocean: a survey on hallucination in large language models.
\newblock {\em arXiv:2309.01219}, 2023.

\bibitem{ye2023cognitive}
Hongbin Ye, Tong Liu, Aijia Zhang, Wei Hua, and Weiqiang Jia.
\newblock Cognitive mirage: A review of hallucinations in large language models.
\newblock {\em arXiv:2309.06794}, 2023.

\bibitem{ho2020denoising}
Jonathan Ho, Ajay Jain, and Pieter Abbeel.
\newblock Denoising diffusion probabilistic models.
\newblock {\em Advances in neural information processing systems}, 33:6840--6851, 2020.

\bibitem{yang2023diffusion}
Ling Yang, Zhilong Zhang, Yang Song, Shenda Hong, Runsheng Xu, Yue Zhao, Wentao Zhang, Bin Cui, and Ming-Hsuan Yang.
\newblock Diffusion models: A comprehensive survey of methods and applications.
\newblock {\em ACM Computing Surveys}, 56(4):1--39, 2023.

\bibitem{rombach2022high}
Robin Rombach, Andreas Blattmann, Dominik Lorenz, Patrick Esser, and Bj{\"o}rn Ommer.
\newblock High-resolution image synthesis with latent diffusion models.
\newblock In {\em Proc. IEEE/CVF conference on computer vision and pattern recognition}, 2022.

\bibitem{yuan2024diffusion}
Xinyu Yuan and Yan Qiao.
\newblock Diffusion-{TS}: Interpretable diffusion for general time series generation.
\newblock {\em arXiv:2403.01742}, 2024.

\bibitem{meijer2024rise}
Caspar Meijer and Lydia~Y Chen.
\newblock The rise of diffusion models in time-series forecasting.
\newblock {\em arXiv:2401.03006}, 2024.

\bibitem{wang2024deep}
Jun Wang, Wenjie Du, Wei Cao, Keli Zhang, Wenjia Wang, Yuxuan Liang, and Qingsong Wen.
\newblock Deep learning for multivariate time series imputation: A survey.
\newblock {\em arXiv:2402.04059}, 2024.

\bibitem{yang2024survey}
Yiyuan Yang, Ming Jin, Haomin Wen, Chaoli Zhang, Yuxuan Liang, Lintao Ma, Yi~Wang, Chenghao Liu, Bin Yang, Zenglin Xu, et~al.
\newblock A survey on diffusion models for time series and spatio-temporal data.
\newblock {\em arXiv:2404.18886}, 2024.

\bibitem{tashiro2021csdi}
Yusuke Tashiro, Jiaming Song, Yang Song, and Stefano Ermon.
\newblock Csdi: Conditional score-based diffusion models for probabilistic time series imputation.
\newblock {\em Advances in Neural Information Processing Systems}, 2021.

\bibitem{xiao2023imputation}
Chunjing Xiao, Zehua Gou, Wenxin Tai, Kunpeng Zhang, and Fan Zhou.
\newblock Imputation-based time-series anomaly detection with conditional weight-incremental diffusion models.
\newblock In {\em Proceedings of the 29th ACM SIGKDD Conference on Knowledge Discovery and Data Mining}, pages 2742--2751, 2023.

\bibitem{chen2023imdiffusion}
Yuhang Chen, Chaoyun Zhang, Minghua Ma, Yudong Liu, Ruomeng Ding, Bowen Li, Shilin He, Saravan Rajmohan, Qingwei Lin, and Dongmei Zhang.
\newblock Imdiffusion: Imputed diffusion models for multivariate time series anomaly detection.
\newblock {\em arXiv:2307.00754}, 2023.

\bibitem{liu2023pristi}
Mingzhe Liu, Han Huang, Hao Feng, Leilei Sun, Bowen Du, and Yanjie Fu.
\newblock Pristi: A conditional diffusion framework for spatiotemporal imputation.
\newblock In {\em 2023 IEEE 39th International Conference on Data Engineering (ICDE)}, 2023.

\bibitem{alcaraz2022diffusion}
Juan Miguel~Lopez Alcaraz and Nils Strodthoff.
\newblock Diffusion-based time series imputation and forecasting with structured state space models.
\newblock {\em arXiv:2208.09399}, 2022.

\bibitem{wang2023observed}
Xu~Wang, Hongbo Zhang, Pengkun Wang, Yudong Zhang, Binwu Wang, Zhengyang Zhou, and Yang Wang.
\newblock An observed value consistent diffusion model for imputing missing values in multivariate time series.
\newblock In {\em Proc. 29th ACM SIGKDD Conference on Knowledge Discovery and Data Mining}, 2023.

\bibitem{zhou2024mtsci}
Jianping Zhou, Junhao Li, Guanjie Zheng, Xinbing Wang, and Chenghu Zhou.
\newblock Mtsci: A conditional diffusion model for multivariate time series consistent imputation.
\newblock {\em arXiv:2408.05740}, 2024.

\bibitem{lugmayr2022repaint}
Andreas Lugmayr, Martin Danelljan, Andres Romero, Fisher Yu, Radu Timofte, and Luc Van~Gool.
\newblock Repaint: Inpainting using denoising diffusion probabilistic models.
\newblock In {\em Proceedings of the IEEE/CVF conference on computer vision and pattern recognition}, pages 11461--11471, 2022.

\bibitem{peng2023check}
Baolin Peng, Michel Galley, Pengcheng He, Hao Cheng, Yujia Xie, Yu~Hu, Qiuyuan Huang, Lars Liden, Zhou Yu, Weizhu Chen, et~al.
\newblock Check your facts and try again: Improving large language models with external knowledge and automated feedback.
\newblock {\em arXiv:2302.12813}, 2023.

\bibitem{shuster2021retrieval}
Kurt Shuster, Spencer Poff, Moya Chen, Douwe Kiela, and Jason Weston.
\newblock Retrieval augmentation reduces hallucination in conversation.
\newblock {\em arXiv:2104.07567}, 2021.

\bibitem{lewis2020retrieval}
Patrick Lewis, Ethan Perez, Aleksandra Piktus, Fabio Petroni, Vladimir Karpukhin, Naman Goyal, Heinrich K{\"u}ttler, Mike Lewis, Wen-tau Yih, Tim Rockt{\"a}schel, et~al.
\newblock Retrieval-augmented generation for knowledge-intensive nlp tasks.
\newblock {\em Advances in Neural Information Processing Systems}, 33:9459--9474, 2020.

\bibitem{chen2023purr}
Anthony Chen, Panupong Pasupat, Sameer Singh, Hongrae Lee, and Kelvin Guu.
\newblock Purr: Efficiently editing language model hallucinations by denoising language model corruptions.
\newblock {\em arXiv:2305.14908}, 2023.

\bibitem{varshney2023stitch}
Neeraj Varshney, Wenlin Yao, Hongming Zhang, Jianshu Chen, and Dong Yu.
\newblock A stitch in time saves nine: Detecting and mitigating hallucinations of llms by validating low-confidence generation.
\newblock {\em arXiv preprint arXiv:2307.03987}, 2023.

\bibitem{mundler2023self}
Niels M{\"u}ndler, Jingxuan He, Slobodan Jenko, and Martin Vechev.
\newblock Self-contradictory hallucinations of large language models.
\newblock {\em arXiv:2305.15852}, 2023.

\bibitem{manakul2023selfcheckgpt}
Potsawee Manakul, Adian Liusie, and Mark~JF Gales.
\newblock Selfcheckgpt: Zero-resource black-box hallucination detection for generative large language models.
\newblock {\em arXiv:2303.08896}, 2023.

\bibitem{elaraby2023halo}
Mohamed Elaraby, Mengyin Lu, Jacob Dunn, Xueying Zhang, Yu~Wang, and Shizhu Liu.
\newblock Halo: Estimation and reduction of hallucinations in open-source weak large language models.
\newblock {\em arXiv:2308.11764}, 2023.

\bibitem{zhang2023sac}
Jiaxin Zhang, Zhuohang Li, Kamalika Das, Bradley~A Malin, and Sricharan Kumar.
\newblock Sac3: Reliable hallucination detection in black-box language models via semantic-aware cross-check consistency.
\newblock {\em arXiv:2311.01740}, 2023.

\bibitem{farquhar2024detecting}
Sebastian Farquhar, Jannik Kossen, Lorenz Kuhn, and Yarin Gal.
\newblock Detecting hallucinations in large language models using semantic entropy.
\newblock {\em Nature}, 630(8017):625--630, 2024.

\bibitem{du2023improving}
Yilun Du, Shuang Li, Antonio Torralba, Joshua~B Tenenbaum, and Igor Mordatch.
\newblock Improving factuality and reasoning in language models through multiagent debate.
\newblock {\em arXiv:2305.14325}, 2023.

\bibitem{chen2023hallucination}
Yuyan Chen, Qiang Fu, Yichen Yuan, Zhihao Wen, Ge~Fan, Dayiheng Liu, Dongmei Zhang, Zhixu Li, and Yanghua Xiao.
\newblock Hallucination detection: Robustly discerning reliable answers in large language models.
\newblock In {\em Proceedings of the 32nd ACM International Conference on Information and Knowledge Management}, pages 245--255, 2023.

\bibitem{pacchiardi2023catch}
Lorenzo Pacchiardi, Alex~J Chan, S{\"o}ren Mindermann, Ilan Moscovitz, Alexa~Y Pan, Yarin Gal, Owain Evans, and Jan Brauner.
\newblock How to catch an ai liar: Lie detection in black-box llms by asking unrelated questions.
\newblock {\em arXiv:2309.15840}, 2023.

\bibitem{mishra2024fine}
Abhika Mishra, Akari Asai, Vidhisha Balachandran, Yizhong Wang, Graham Neubig, Yulia Tsvetkov, and Hannaneh Hajishirzi.
\newblock Fine-grained hallucination detection and editing for language models.
\newblock {\em arXiv preprint arXiv:2401.06855}, 2024.

\bibitem{zha2023alignscore}
Yuheng Zha, Yichi Yang, Ruichen Li, and Zhiting Hu.
\newblock Alignscore: Evaluating factual consistency with a unified alignment function.
\newblock {\em arXiv preprint arXiv:2305.16739}, 2023.

\bibitem{su2024unsupervised}
Weihang Su, Changyue Wang, Qingyao Ai, Yiran Hu, Zhijing Wu, Yujia Zhou, and Yiqun Liu.
\newblock Unsupervised real-time hallucination detection based on the internal states of large language models.
\newblock {\em arXiv preprint arXiv:2403.06448}, 2024.

\bibitem{gu2024anah}
Yuzhe Gu, Ziwei Ji, Wenwei Zhang, Chengqi Lyu, Dahua Lin, and Kai Chen.
\newblock Anah-v2: Scaling analytical hallucination annotation of large language models.
\newblock {\em arXiv preprint arXiv:2407.04693}, 2024.

\bibitem{gu2025mask}
Yuzhe Gu, Wenwei Zhang, Chengqi Lyu, Dahua Lin, and Kai Chen.
\newblock Mask-dpo: Generalizable fine-grained factuality alignment of llms.
\newblock {\em arXiv preprint arXiv:2503.02846}, 2025.

\bibitem{tian2023fine}
Katherine Tian, Eric Mitchell, Huaxiu Yao, Christopher~D Manning, and Chelsea Finn.
\newblock Fine-tuning language models for factuality.
\newblock In {\em The Twelfth International Conference on Learning Representations}, 2023.

\bibitem{lin2024flame}
Sheng-Chieh Lin, Luyu Gao, Barlas Oguz, Wenhan Xiong, Jimmy Lin, Scott Yih, and Xilun Chen.
\newblock Flame: Factuality-aware alignment for large language models.
\newblock {\em Advances in Neural Information Processing Systems}, 37:115588--115614, 2024.

\bibitem{zhang2024self}
Xiaoying Zhang, Baolin Peng, Ye~Tian, Jingyan Zhou, Lifeng Jin, Linfeng Song, Haitao Mi, and Helen Meng.
\newblock Self-alignment for factuality: Mitigating hallucinations in llms via self-evaluation.
\newblock {\em arXiv preprint arXiv:2402.09267}, 2024.

\bibitem{chen2024grath}
Weixin Chen, Dawn Song, and Bo~Li.
\newblock Grath: Gradual self-truthifying for large language models.
\newblock {\em arXiv preprint arXiv:2401.12292}, 2024.

\bibitem{zhang2023language}
Muru Zhang, Ofir Press, William Merrill, Alisa Liu, and Noah~A Smith.
\newblock How language model hallucinations can snowball.
\newblock {\em arXiv:2305.13534}, 2023.

\bibitem{gallifant2024peer}
Jack Gallifant, Amelia Fiske, Yulia~A Levites~Strekalova, Juan~S Osorio-Valencia, Rachael Parke, Rogers Mwavu, Nicole Martinez, Judy~Wawira Gichoya, Marzyeh Ghassemi, Dina Demner-Fushman, et~al.
\newblock Peer review of {GPT-4} technical report and systems card.
\newblock {\em PLOS Digital Health}, 3(1):e0000417, 2024.

\bibitem{yang2024generalized}
Jingkang Yang, Kaiyang Zhou, Yixuan Li, and Ziwei Liu.
\newblock Generalized out-of-distribution detection: A survey.
\newblock {\em International Journal of Computer Vision}, 132(12):5635--5662, 2024.

\bibitem{zamanzadeh2024deep}
Zahra Zamanzadeh~Darban, Geoffrey~I Webb, Shirui Pan, Charu Aggarwal, and Mahsa Salehi.
\newblock Deep learning for time series anomaly detection: A survey.
\newblock {\em ACM Computing Surveys}, 57(1):1--42, 2024.

\bibitem{aithal2024understanding}
Sumukh~K Aithal, Pratyush Maini, Zachary~C Lipton, and J~Zico Kolter.
\newblock Understanding hallucinations in diffusion models through mode interpolation.
\newblock {\em arXiv:2406.09358}, 2024.

\bibitem{ecldataset}
Artur Trindade.
\newblock {ElectricityLoadDiagrams20112014}.
\newblock UCI Machine Learning Repository, 2015.
\newblock {DOI}: https://doi.org/10.24432/C58C86.

\bibitem{wthdataset}
{Max Planck Institute for Biogeochemistry}.
\newblock Weather data.
\newblock \url{https://www.bgc-jena.mpg.de/wetter/}, 2024.
\newblock Accessed: 2025-01-16.

\bibitem{trafficdataset}
{California Department of Transportation}.
\newblock Performance measurement system (pems).
\newblock \url{http://pems.dot.ca.gov/}, 2024.
\newblock Accessed: 2025-01-16.

\bibitem{illnessdataset}
{Centers for Disease Control and Prevention}.
\newblock Fluview: Flu activity \& surveillance.
\newblock \url{https://gis.cdc.gov/grasp/fluview/fluportaldashboard.html}, 2024.
\newblock Accessed: 2025-01-16.

\bibitem{zhou2021informer}
Haoyi Zhou, Shanghang Zhang, Jieqi Peng, Shuai Zhang, Jianxin Li, Hui Xiong, and Wancai Zhang.
\newblock Informer: Beyond efficient transformer for long sequence time-series forecasting.
\newblock In {\em Proc. AAAI conference on Artificial Intelligence}, 2021.

\bibitem{nie2022time}
Yuqi Nie, Nam~H Nguyen, Phanwadee Sinthong, and Jayant Kalagnanam.
\newblock A time series is worth 64 words: Long-term forecasting with transformers.
\newblock {\em arXiv:2211.14730}, 2022.

\bibitem{diederik2014adam}
P~Kingma Diederik.
\newblock Adam: A method for stochastic optimization.
\newblock {\em arXiv:1412.6980}, 2014.

\bibitem{smith2019super}
Leslie~N Smith and Nicholay Topin.
\newblock Super-convergence: Very fast training of neural networks using large learning rates.
\newblock In {\em Artificial intelligence and machine learning for multi-domain operations applications}, volume 11006, pages 369--386. SPIE, 2019.

\end{thebibliography}
\bibliographystyle{unsrt}


\clearpage
\appendix

\section{Other Metrics}
\label{appendix:othermetrics}

\begin{figure}[h]
    \centering
    \includegraphics[width=0.6\linewidth]{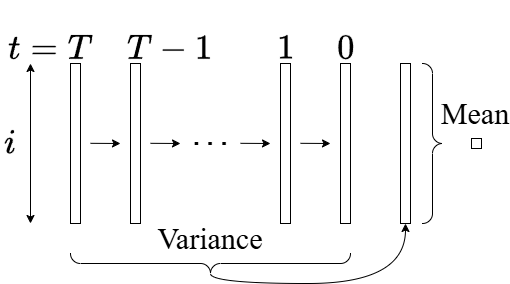}
    \caption{Schematic showing the computation of the trajectory variance (TV) metric. The variance is taken across the diffusion time-step $t$ and the mean is taken across the data dimension $i$.}
    \label{fig:ts_computing}
\end{figure}

It has been shown on a computer vision and toy Gaussian dataset that a measure of hallucination can be extracted from unconditional diffusion models during the generation process \cite{aithal2024understanding}. This measure will be referred to as the the trajectory variance (TV), which is the variance of the predicted mean with respect to the diffusion time-step. The predicted mean at each diffusion time-step (Eq. \ref{eq:mu}) is obtained from the generation process and will be written as $\mu_{i,t}$, where $i$ indexes the data dimension and $t$ indexes the diffusion time-step. The TV metric is calculated as
\begin{equation}
    M_{\mathrm{TV}} = \mathrm{Mean}_i\bigg(\mathrm{Var}_t(\mu_{i,t}) \bigg),
\end{equation}
where variance is taken across the diffusion time-step and  mean across the data dimension. A schematic example of this is shown in Fig.~\ref{fig:ts_computing}. This measures the variation in the trajectory of the variables during the diffusion process. 

TV however only applies to unconditional generation and does not apply to the prompt-response framework using imputation. This is because in the prompt-response framework, the subset of the data dimension that is used for the prompt is not unconditionally generated. Three modifications to the TV metric that address this are proposed. These are response trajectory spread (RTS), prompt trajectory spread (PTS) and combined trajectory spread (CTS) metrics. We also propose two additional metrics that use the magnitude of the noise returned by the diffusion model as a metric to detect hallucination. These are prompt error (PE) and combined error (CE). The combined error is the metric presented in the main text as this is the most effective metric and the other metrics fail at detecting relational hallucination.

\paragraph{Response Trajectory Spread (RTS)}

Since TV is computed for unconditional generation, the simplest generalisation to the prompt-response framework is to compute this metric for the response only, as this is the part which is generated in a similar manner. Instead of using the variance however, this work uses the standard deviation since it is simpler and is more interpretable. The response trajectory spread (RTS) can be computed as
\begin{equation}
    M_{\mathrm{RTS}} = \mathrm{Mean}_i\bigg(\mathrm{Std}_t(\mu_{i, t}) \bigg), \;\forall i \in \mathcal{I}_{\mathrm{r}},
\end{equation}
where standard deviation is taken across the diffusion time-step, and mean across the data dimension. This is illustrated in Fig.~\ref{fig:ts_computing} but with  standard deviation instead of  variance.

\paragraph{Prompt Trajectory Spread (PTS)} 
As we are using RePaint \cite{lugmayr2022repaint} to condition the diffusion model, all predicted means of the prompt are clamped to the values provided by the prompt. The values provided by the prompt can therefore be used as the mean that is required to compute the standard deviation. The prompt trajectory spread (PTS) can be computed as
\begin{equation}
    M_{\mathrm{PTS}} = \mathrm{Mean}_i\bigg(\mathrm{RMSE}_t(\mu_{i, t}, x_{i}) \bigg), \;\forall i \in \mathcal{I}_{\mathrm{p}},
\end{equation}
where the root mean square error is taken across the diffusion time-step and the mean across the data dimension. 

\paragraph{Combined Trajectory Spread (CTS)} 
The final output from the model $\hat{x}_{i}$ which combines both the prompt and the response can also be used to compute the trajectory spread. The full diffusion process can be computed one more time by setting the prompt as $x_{i} = \hat{x}_{i}, \forall i \in \mathcal{I}$. The combined trajectory spread (CTS) can then be computed for this as
\begin{equation}
    M_{\mathrm{CTS}} = \mathrm{Mean}_i\bigg(\mathrm{RMSE}_t(\mu_{i, t}, \hat{x}_{i} ) \bigg), \;\forall i \in \mathcal{I},
\end{equation}
where the root mean square error $RMSE_t(\mu_{i,t},x_{i}) = \sqrt{\frac{1}{T}\sum_{t=1}^{T}(\mu_{i,t} - x_{i})^{2}} $ is taken across the diffusion time-step and the mean is taken across the data dimension. Since the full diffusion process has to be computed completely an additional time, this metric is computationally expensive. CTS is like PTS but includes both the prompt and response.

\paragraph{Prompt Error (PE)}

The previous metrics are all based on the trajectory variance \cite{aithal2024understanding}. This work proposes two additional simple metrics based on the reconstruction error of the final output of the model. The first considers the reconstruction error of the output with respect to the prompt. Only indices $i \in \mathcal{I}_{\mathrm{P}}$ are used since as are the only values where ground-truth is available through the values provided by the prompt. PE can be computed as 
\begin{equation}
    M_{\mathrm{PE}} = \mathrm{RMSE}_i(\hat{x}_{i}, x_{i}), \;\forall i \in \mathcal{I}_{\mathrm{p}},
\end{equation}
where the  RMSE is taken across the data dimension.

\paragraph{Combined Error (CE)}
The PE metric can be extended to also include the response indices $\mathcal{I}_{\mathrm{R}}$ in the same way as the CTS metric, which leads to the CE metric of Eqn.~\ref{eqn:def-ce}.

\subsection{Sensitivity to Relational Hallucination}
\begin{figure}[H]
    \centering
    \includegraphics[width=1.0\linewidth]{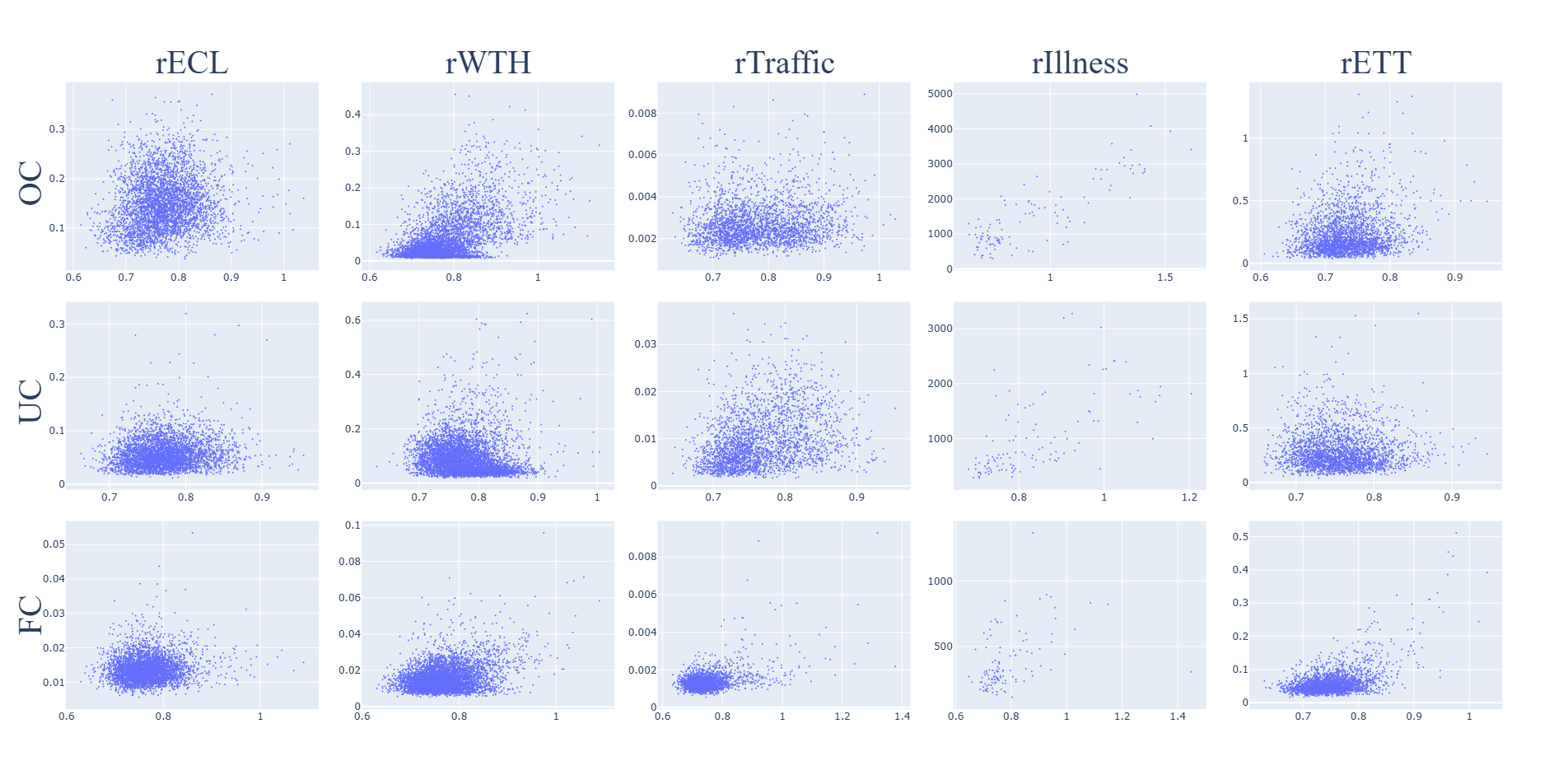}
    \caption{Scatter plot showing the relationship between the RTS metric ($x$-axis) and the ground-truth relational error ($y$-axis) on the test set. The subplots are aligned by dataset (column) and task (row). The axis limits are the same within each dataset (column).}
    \label{fig:rts_test_all}
\end{figure}

\paragraph{RTS} 
The sensitivity of the RTS metric to the relational error on the test set for each task and dataset is shown in Fig.~\ref{fig:rts_test_all}. The metric is not sensitive to the relational error.

\begin{figure}[H]
    \centering
    \includegraphics[width=1.0\linewidth]{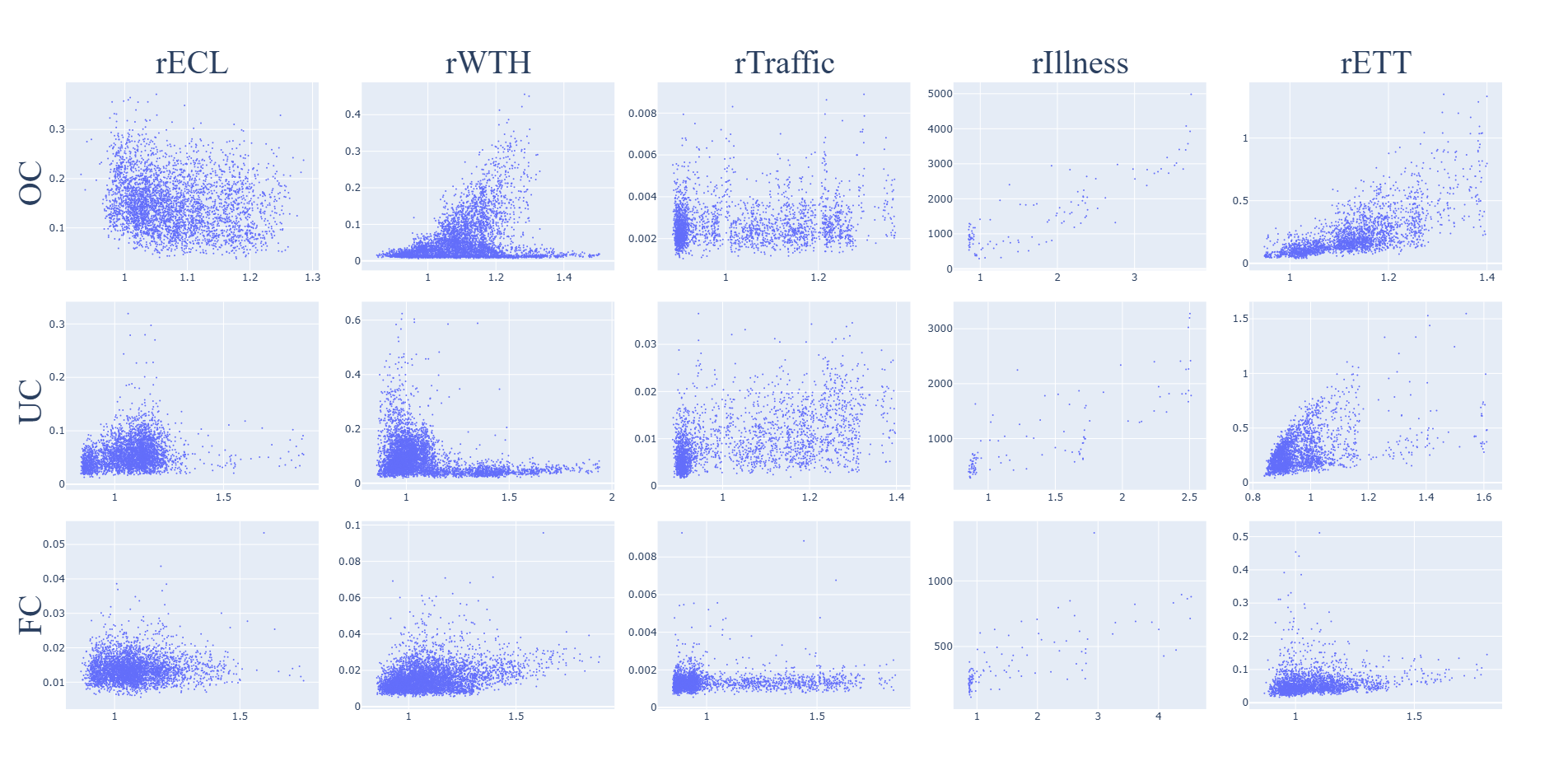}
    \caption{Scatter plot showing the relationship between the PTS metric ($x$-axis) and the ground-truth relational error ($y$-axis) on the test set. The subplots are aligned by dataset (column) and task (row). The axis limits are the same within each dataset (column).}
    \label{fig:pts_test_all}
\end{figure}

\paragraph{PTS}
The sensitivity of the PTS metric to the relational error on the test set for each task and dataset is shown in Fig.~\ref{fig:pts_test_all}. The metric is not sensitive to the relational error.

\begin{figure}[H]
    \centering
    \includegraphics[width=1.0\linewidth]{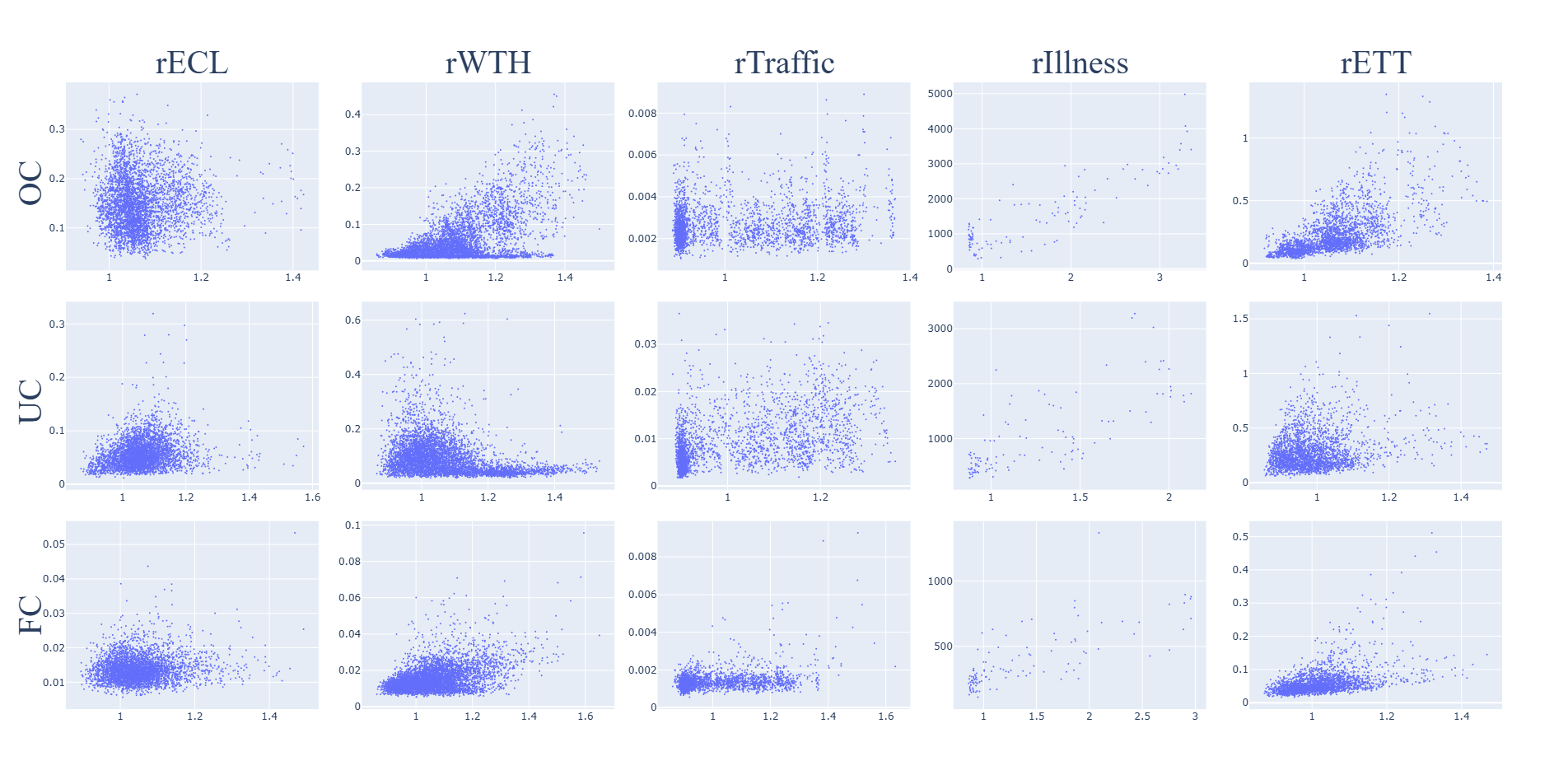}
    \caption{Scatter plot showing the relationship between the CTS metric ($x$-axis) and the ground-truth relational error ($y$-axis) on the test set. The subplots are aligned by dataset (column) and task (row). The axis limits are the same within each dataset (column).}
    \label{fig:cts_test_all}
\end{figure}

\paragraph{CTS}
The sensitivity of the CTS metric to the relational error on the test set for each task and dataset is shown in Fig.~\ref{fig:cts_test_all}. The metric is not sensitive to the relational error.

\begin{figure}[H]
    \centering
    \includegraphics[width=1.0\linewidth]{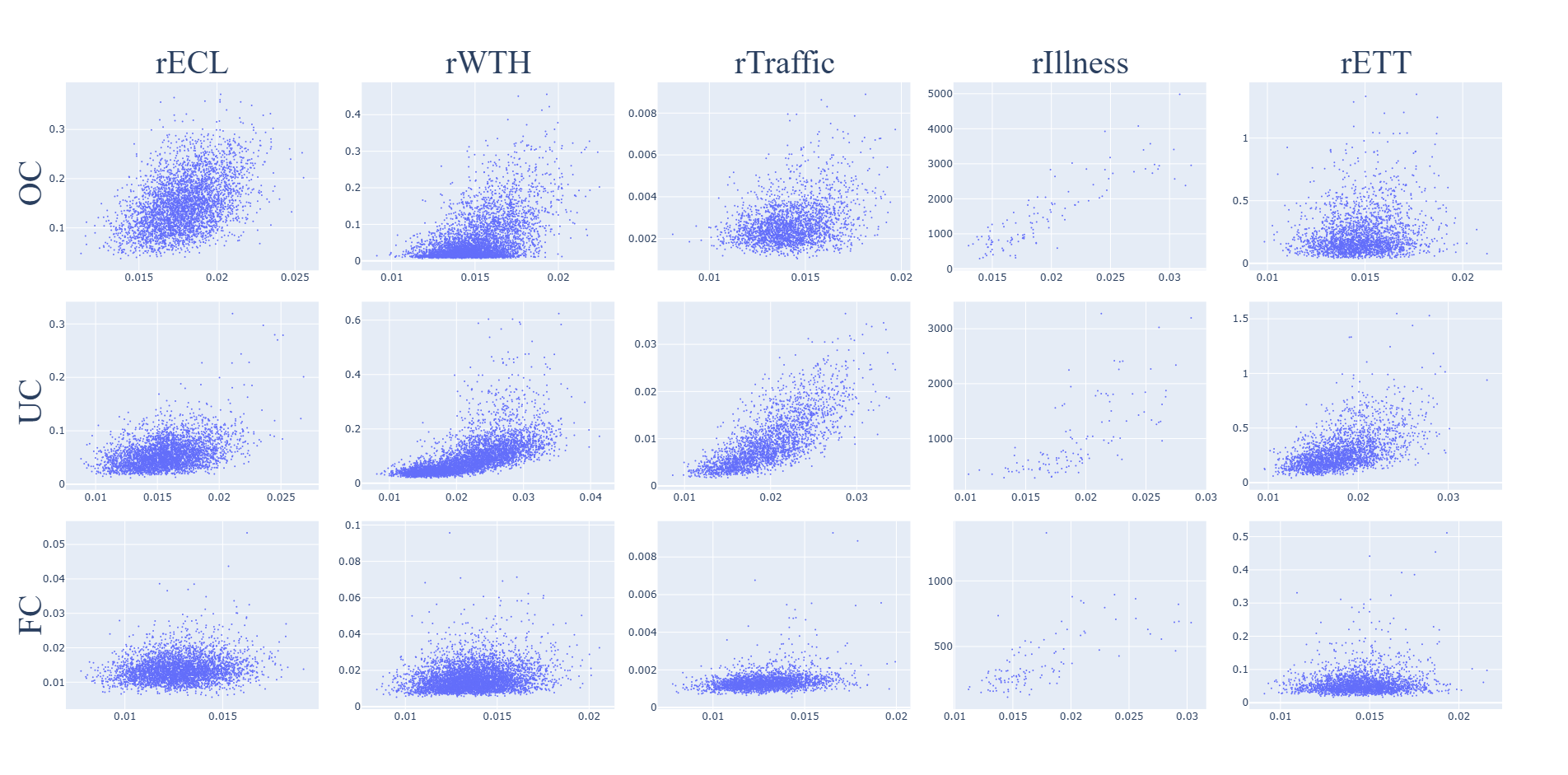}
    \caption{Scatter plot showing the relationship between the PE metric ($x$-axis) and the ground-truth relational error ($y$-axis) on the test set. The subplots are aligned by dataset (column) and task (row). The axis limits are the same within each dataset (column).}
    \label{fig:pe_test_all}
\end{figure}

\begin{figure}[H]
    \centering
    \includegraphics[width=1.0\linewidth]{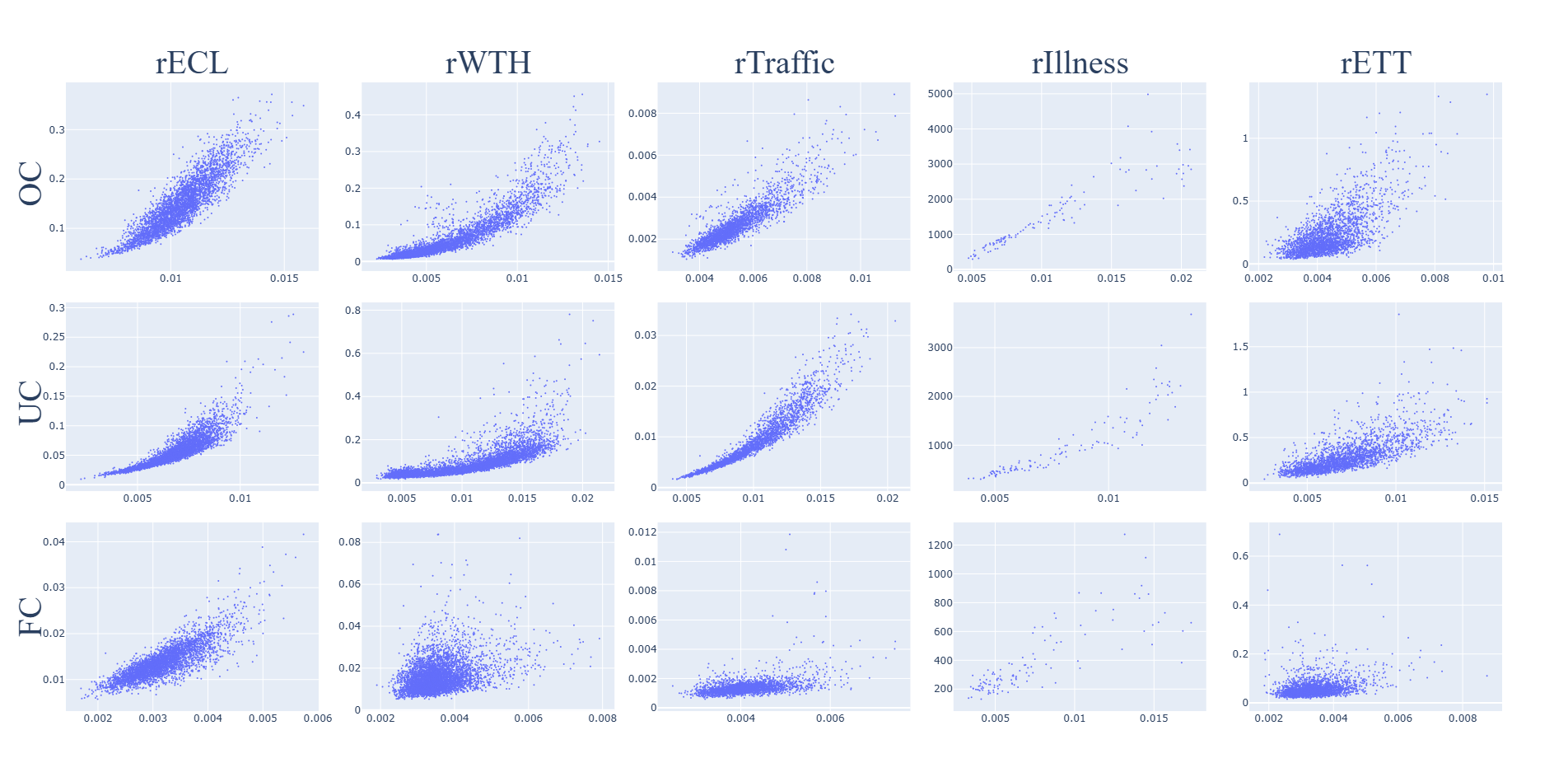}
    \caption{Scatter plot showing the relationship between the CE metric ($x$-axis) and the ground-truth relational error ($y$-axis) on the test set. The subplots are aligned by dataset (column) and task (row). The axis limits are the same within each dataset (column).}
    \label{fig:ce_test_all}
\end{figure}

\paragraph{PE}
Sensitivity of PE to relational error on the test set for each task and dataset is shown in Fig.~\ref{fig:pe_test_all}. PE is not as robustly and consistently sensitive to the relational error as the CE metric, which is shown in Fig. \ref{fig:ce_test_all}.  This may be because PE only includes the prompt, and since relational hallucination is the inconsistency of a prompt-response pair, it is expected that a metric including both the prompt and response such as CE  would perform better.

\paragraph{CE}
Sensitivity of CE to relational error on the test set for each task and dataset is shown in Fig.~\ref{fig:ce_test_all}. The CE metric is sensitive to the relational error.

\section{Licenses for Existing Assets}
\label{appendix:licenses}

\subsection{Datasets}
The datasets are commonly used MVTS time-series datasets and can be accessed from the Autoformer repository (https://github.com/thuml/Autoformer) which is under MIT license.

\begin{itemize}
    \item ECL - CC BY 4.0
    \item WTH - N/A
    \item Traffic - CC BY 4.0
    \item Illness - N/A
    \item ETT - CC BY-ND 4.0
\end{itemize}

\subsection{Models}
\begin{itemize}
    \item MOMENT - MIT (https://github.com/moment-timeseries-foundation-model/moment)
    \item TIMER - MIT (https://github.com/thuml/Large-Time-Series-Model)
\end{itemize}


\end{document}